\documentclass[conference]{IEEEtran}
\IEEEoverridecommandlockouts
\usepackage{multirow}
\usepackage{booktabs} % For better table lines

\usepackage{cite}
\usepackage{amsmath,amssymb,amsfonts}
\usepackage{algorithm, algorithmic}
\usepackage{graphicx}
\usepackage{textcomp}
\usepackage{xcolor}
\usepackage{adjustbox}
\usepackage{listings}

\def\BibTeX{{\rm B\kern-.05em{\sc i\kern-.025em b}\kern-.08em
    T\kern-.1667em\lower.7ex\hbox{E}\kern-.125emX}}
\begin{document}

\iffalse{
\title{FTP: First Token Probability-Guided Context Adjustment in LLM RAG for MCQA}
}\fi
\title{First Token Probability Guided RAG for Telecom Question Answering}

\author{%
    \IEEEauthorblockN{Tingwei Chen\IEEEauthorrefmark{1}, Jiayi Chen\IEEEauthorrefmark{1}, Zijian Zhao\IEEEauthorrefmark{1}\IEEEauthorrefmark{3}, Haolong Chen\IEEEauthorrefmark{1}\IEEEauthorrefmark{2}, Liang Zhang\IEEEauthorrefmark{1}, Guangxu Zhu\IEEEauthorrefmark{1}  \thanks{Corresponding author: Guangxu Zhu} }%
    \IEEEauthorblockA{\IEEEauthorrefmark{1} Shenzhen Research Institute of Big Data, Shenzhen, China \\ % 
    \IEEEauthorrefmark{2} School of Science and Engineering, The Chinese University of Hong Kong (Shenzhen), Shenzhen, China\\ %
    \IEEEauthorrefmark{3} School of Computer Science and Engineering, Sun Yat-sen University, Guangzhou, China\\ %
    }
    % Email:\{tingweichen,224010118,haolongchen1\}@link.cuhk.edu.cn,\{zhaozj28\}@mail2.sysu.edu.cn,\{zhangliang,gxzhu\}@sribd.cn
    %
}

%First Token first token probability-Guided Context Adjustment in Domain-Specific RAG for MCQA

% \and
% \and
% \and
% \and
% \and

\maketitle

\begin{abstract}
Large Language Models (LLMs) have garnered significant attention for their impressive general-purpose capabilities. For applications requiring intricate domain knowledge, Retrieval-Augmented Generation (RAG) has shown a distinct advantage in incorporating domain-specific information into LLMs. However, existing RAG research has not fully addressed the challenges of Multiple Choice Question Answering (MCQA) in telecommunications, particularly in terms of retrieval quality and mitigating hallucinations. To tackle these challenges, we propose a novel first token probability guided RAG framework. This framework leverages confidence scores to optimize key hyperparameters, such as chunk number and chunk window size, while dynamically adjusting the context. Our method starts by retrieving the most relevant chunks and generates a single token as the potential answer. The probabilities of all options are then normalized to serve as confidence scores, which guide the dynamic adjustment of the context. By iteratively optimizing the hyperparameters based on these confidence scores, we can continuously improve RAG performance. We conducted experiments to validate the effectiveness of our framework, demonstrating its potential to enhance accuracy in domain-specific MCQA tasks.

% This method uses a prompt template that directs the model to generate only a single token as the answer, and it normalizes the probabilities of the choices in MCQA to serve as confidence scores, providing a standard for evaluating the LLMs' output. 

% Based on the confidence scores, we assess the hyperparameter and iteratively optimize it to achieve better performance. This iterative hyperparameter optimization process allows for continuous improvement in RAG performance. We conducted experiments to demonstrate the effectiveness of our RAG framework.
\end{abstract}


\section{Introduction}
Large Language Models (LLMs) have garnered significant attention due to their impressive general-purpose capabilities~\cite{zhao2023survey, zhou2024large,chen2024overview}. These generative language models can understand and process user inputs, comprehend complex knowledge, and generate high-quality answers, demonstrating extensive understanding of general human knowledge. As LLMs have shown excellent general capabilities, they have been applied not only in natural language, but also in fields like code~\cite{ouyang2023llm}, mathematics~\cite{wu2024mathchat}, and music~\cite{liang2024pianobart}.

For specific applications with intricate domain knowledge, LLMs have a potential shortfall in the precision and depth of knowledge and thus limit their ability to fully meet the nuanced demands. To inject domain-specific knowledge into LLMs, Retrieval-Augmented Generation (RAG) shows a significant advantage in that it does not require the huge computational resources and can efficiently extract useful information from a large external knowledge base, which can even be updated at any time without incurring additional costs~\cite{li2022survey}. RAG enhances the capability of LLMs to address specific industry challenges by leveraging external knowledge bases during the generation process. This method involves two main components: a retriever and a generator. The retriever component searches a large corpus of domain-specific chunks to find relevant information, while the generator component uses this information to produce contextually appropriate responses. The core of RAG is retrieving the most relevant information efficiently and accurately and transferring it as an efficient prompt to make it understandable by LLMs, all of which have been the subject of extensive research~\cite{karpukhin2020dense,edge2024local,yu2024rankrag}.

In the field of telecommunications, the rapid advancement and complexity of technologies necessitate a deep understanding of communication protocols. Professionals in this domain need to meticulously retrieve and study these protocols to grasp the intricate logical mechanisms and technical details that govern the entire communication process. With the continuous evolution of 5G, 6G, and forthcoming generations of communication protocols, the corpus of knowledge in this domain is ever-expanding. This growth imposes an increasing burden on professionals, as the time and effort required for effective information retrieval escalate. Traditional methods of manual retrieval and study in~\cite{karpukhin2020dense,edge2024local,yu2024rankrag} are becoming increasingly impractical and inefficient in handling the vast amount of information.

Recently, there exist few research works focusing on the domain-specific RAG for the telecommunication applications. By combining the communication, understanding, and thinking capabilities of LLMs with the expertise from knowledge bases, AI agents can play a useful role in many scenarios, such as training employees, answering professional questions from clients, and generating professional reports. Especially as telecommunications data, such as protocols and standards, are continuously updated, RAG can efficiently overcome this without requiring the model to be retrained. For example, Bornea et al. propose Telco-RAG~\cite{bornea2024telco}, a universal framework to process telecommunications standards using components like routers to augment queries with technical glossaries and neural networks to locate information in the knowledge base. This approach significantly enhances the model's ability to provide detailed and accurate answers compared to relying solely on the information encoded in the LLM's training data. Additionally, Yilma et al.~\cite{yilma2024telecomrag} propose TelecomRAG, which enhances traditional RAG by adding role-playing in the prompt and verifying and optimizing the output for appropriateness. While there is still room for improvement in the retrieval performance of existing RAG implementations. One of the key challenges is ensuring that the retriever accurately identifies the most relevant documents from a potentially vast and diverse corpus. Additionally, the generator must effectively integrate and contextualize the retrieved information to produce coherent and informative responses.

However, existing RAG research has not fully addressed the specific challenges of MCQA tasks~\cite{wang2024look, robinson2023leveraginglargelanguagemodels}, particularly within the telecommunications domain. Instead of generating answers directly, the model needs to select the correct answer from a set of multiple choices based on the given question. Only one recent work, Telco-RAG~\cite{bornea2024telco}, has shown potential in handling telecommunications MCQA, which typically relies on models with large parameter counts. When smaller models are used~\cite{ahmed2024linguistic}, the challenge of effectively matching the correct options becomes more pronounced. It is difficult to retrieve high-quality information, and smaller models often struggle to capture the nuanced domain knowledge required to answer questions accurately. Furthermore, when using language models for question answering, the tendency of these models to generate plausible but incorrect responses—known as "hallucinations"—introduces an additional risk in this accuracy-critical domain. When LLMs select an answer, they generate multi-token choice descriptions directly without a clear indication of confidence in their decision, making it difficult to trust and understand the reasoning behind their responses.

To address the hallucination issue, we propose a novel first token probability method to measure the confidence of the model's decisions. Specifically, we design a prompt template that directs the model to generate only a single token as the answer. This approach reduces computational costs and minimizes the likelihood of producing irrelevant tokens. We then normalize the probabilities of the choices in MCQA to serve as confidence scores, providing a standard for evaluating the LLMs' output. By setting a threshold, we can filter out low-confidence answers, thereby reducing the potential for hallucinations. To validate this method, our experiments show that the probability of the generated token is closely linked to its likelihood of being the correct option, offering a more reliable basis for answer selection.

To address the quality issues during retrieval, we propose a novel first token probability guided RAG framework that incorporates global context adjustment with optimized hyperparameters. Parameters like chunk size are crucial in determining the quality of retrieved chunks. Our method tailors hyperparameters to individual questions, increasing the likelihood of capturing the varied relationships between chunks and answers across different questions. Specifically, we begin by setting an initial hyperparameter configuration to retrieve the most relevant chunks. We then apply the first token probability method to evaluate these chunks and the question, generating a judgment score. If the score is unsatisfactory, we adjust the hyperparameters to explore better potential outcomes. This iterative hyperparameter optimization process allows for continuous improvement of RAG performance.

In summary, we have the following contributions
\begin{itemize}
\item We proposed a novel first token probability method to measure decision confidence and verified its consistency with the correct option in MCQA.
\item We introduced a first token probability guided RAG framework that employs a global search for optimized hyperparameters, enabling continuous performance improvement of RAG.
\item We conducted experiments to demonstrate the effectiveness of our RAG framework.
\end{itemize}

\section{Methodology}
In this section, we outline the key methods and approaches employed to tackle the challenges identified in the context of MCQA within the telecommunications domain. Our approach combines prompt engineering and RAG to enhance model performance by effectively leveraging domain-specific knowledge. Additionally, we utilized first token probability as a confidence indicator to guide the hyperparameter optimization within the RAG framework.

\begin{figure*}[h]
    \centering
    \includegraphics[width=1\textwidth]{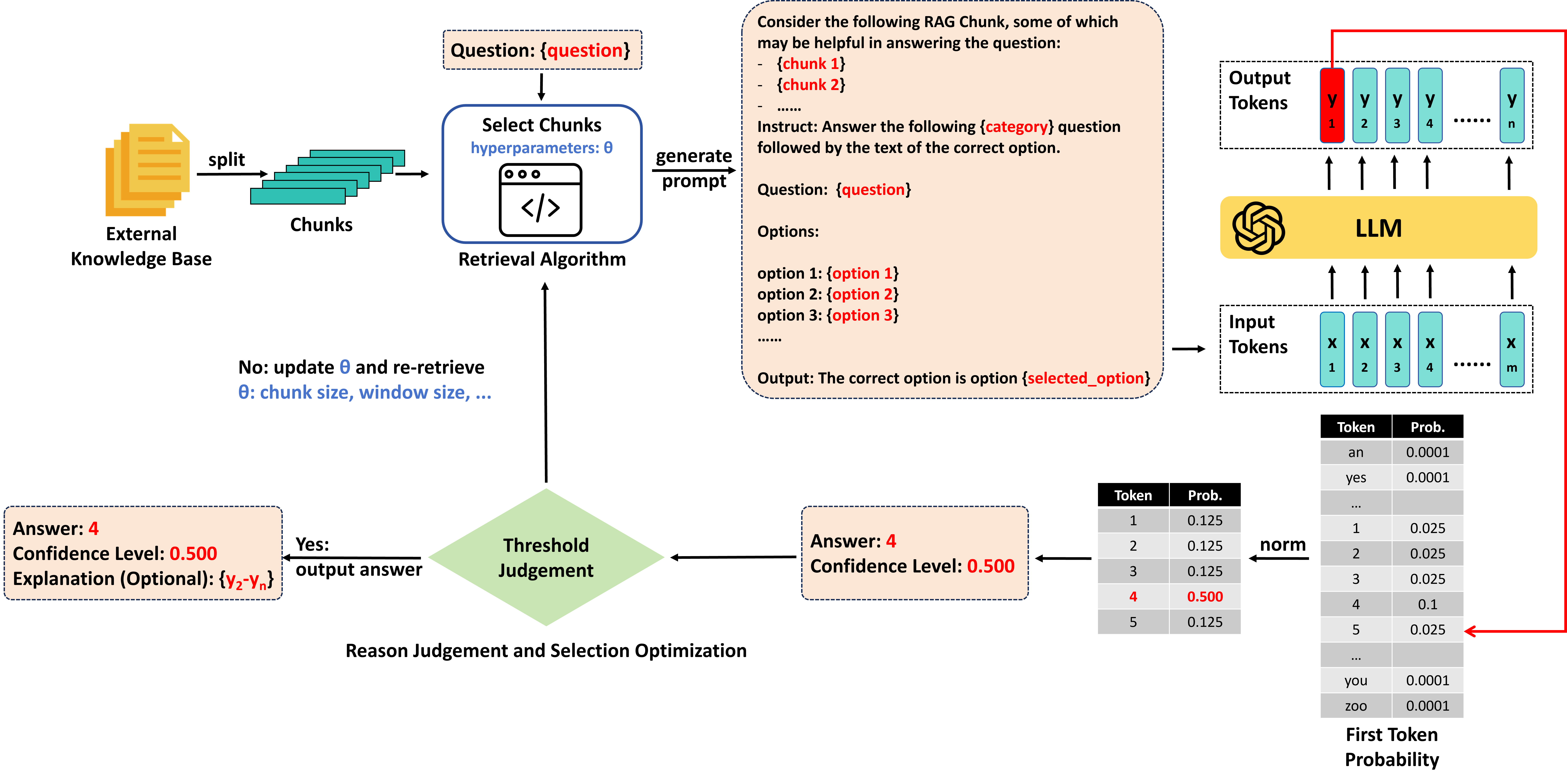}
    \caption{Illustration of the proposed framework of first token probability Guided RAG}
    \label{fig:Main}
\end{figure*}

\subsection{Language models}

Phi-2 is a Small Language Model (SLM) with 2.7 billion parameters, designed for efficient performance in specialized tasks. Despite its compact size, Phi-2 excels in reasoning and language comprehension, comparable to much larger models.~\cite{piovesan2024telecom}

\subsection{Domain-Specific Knowledge Embedding via RAG}
To further improve the accuracy and relevance of responses, we integrated domain-specific knowledge into our system using RAG. This technique allows the model to draw upon a vast external knowledge base, ensuring that the information it retrieves is both accurate and contextually appropriate for the telecommunications domain. The following subsections detail the key components of our RAG implementation, focusing on the challenges associated with chunk size, chunk window size, embedding models, indexing strategy, and top-k chunk selection.
% Retrieval-Augmented Generation (RAG) is a technique that combines the strengths of retrieval-based methods with the generative capabilities of large language models. In RAG, relevant documents or text chunks are retrieved from a corpus based on a query, and these retrieved texts are then used to augment the input to Phi-2.

\subsubsection{Chunk Size}
As RAG methods work by retrieving chunks from a knowledge base, the quality of the split chunks directly influences the model performance. To maintain sentence integrity during chunking, we utilized the SentenceSplitter class from LlamaIndex\footnote{https://github.com/jerryjliu/llama\_index}, which splits text into dependent sentences. However, as the input length of embedding models and LLMs are all limited, we must restrict the maximum length of chunks. As a result, when split sentences are longer than the threshold, we need to further split them into smaller chunks. Additionally, due to the limited precision of sentence splitter models, we noticed that they sometimes split sentences in an incorrect way. For example, a period ('.') may not always represent the end of a sentence, but the model may still split the sentence at that point. These two issues can both lead to incoherent sentences, which may further harm the model's performance intuitively.

\subsubsection{Chunk Window Size}
In RAG, we hypothesized that the chunk might split complete sentences into semantically incomplete fragments, degrading the retrieval effectiveness. To address this, we explored a solution involving window size selection. After retrieving a chunk, we revert to the source text in the knowledge base and construct a single, comprehensive sentence. This is achieved by centering on the key text corresponding to the chunk and merging its preceding and succeeding context. This integrated sentence serves as the external knowledge input to LLMs. The extent of the surrounding context is determined by a variable termed 'Chunk Window Size'. By employing this method, we can ensure that the sentences containing key text remain intact while also increasing the input of knowledge relevant to the query.

% However, we did not observe any performance improvement. This lack of enhancement, in turn, seems to refute the assumption that sentence fragmentation by the chunk negatively affects the accuracy of multiple-choice questions.

\subsubsection{Embedding Models}
Matching chunks is a critical component of the RAG process, necessitating the translation of textual data into numerical feature vectors via an embedding model. This model is tasked with converting text segments of varying lengths into a standardized format of fixed-length numerical vectors. It must also ensure that vectors representing semantically similar content are positioned closely within the feature space to enhance the chunk matching process.

\subsubsection{Indexing Strategy}
After converting external textual knowledge into high-dimensional vector representations via an embedding model, we need to search for vectors that are most closely related to the content of the question within the knowledge base. Vector indexing techniques are crucial for enhancing the efficiency of nearest neighbor searches in high-dimensional spaces, particularly within the domains of machine learning and data mining, where they facilitate the rapid retrieval of data points most similar to a given query vector.

% \subsubsection{Context Length}
% The phi-2 model has a context length limit of 2048 tokens. To prevent token overflow, we limited the chunk size used in processing to 2048 tokens. Given that each chunk averages 44 tokens, this setup allows for a maximum of approximately 45 chunks. However, since the length of retrieved chunks can vary, we carefully manage the total token count, including RAG tokens, to ensure it does not exceed 2048 tokens. 

\subsubsection{Top-K Chunk Number}
Selecting the appropriate external knowledge for a domain-specific MCQA task is essential for enhancing model performance. As previously discussed, determining the chunk size and the matching strategy are critical steps, with the number of knowledge chunks emerging as a pivotal hyperparameter. Contrary to what might be expected, our experiments have shown that incrementally increasing the top-k parameter does not guarantee a consistent improvement in performance. A potential explanation is that a larger k value introduces more irrelevant chunks, potentially causing knowledge interference within the LLM.

\subsection{Prompt Engineering}
Prompt engineering may provide a promising approach to address some of the challenges mentioned earlier in MCQA, particularly in overcoming inconsistencies in answer selection and improving overall accuracy. As illustrated in Fig. \ref{fig:Main}, instead of generating full-text answers, an efficient strategy involves directing the model to output a single token corresponding to the selected answer. By evaluating the logits of the first token generated for each option and selecting the one with the highest probability, we streamline the selection process and reduce inference time. This method avoids the inherent complexities of aligning free-form text outputs with predefined options, significantly improving the accuracy of answer selection. %As illustrated in Fig. \ref{fig:QA}, this single-token approach achieved an accuracy of 51.64\%, markedly outperforming methods that rely on generating full-text responses. This result emphasizes the importance of structured prompt design in enhancing the performance of MCQA systems.

\subsection{First Token Probability}

When a model generates tokens, it leverages prior knowledge from the context to produce a probability distribution across the entire vocabulary for the next token. To mitigate uncertainty in generating the correct option token, one approach is to select the token with the highest probability. However, when the contextual information is insufficient to allow the model to generate a correct answer, the token with the highest probability may not have a overwhelming probability.

The experiments presented in the following section show that there is a positive correlation between first token probability and the accuracy of predictions in MCQA tasks. This means that as the first token probability increases, the likelihood that the model's selected answer is correct also increases. Intuitively, this relationship arises because a higher first token probability indicates that the model is more confident in its prediction, which often translates to higher accuracy.

Given this relationship, first token probability can be used as a reliable metric to guide context adjustment. By monitoring this probability, we can dynamically decide whether the context information is sufficient or if additional chunks need to be retrieved to improve confidence, and thus, accuracy.

\subsection{Hyperparameters Optimization}
Hyperparameters Optimization is crucial for enhancing the performance of a RAG system. By using first token probability as a metric, we can assess the effectiveness of the chosen hyperparameters in answering a given question. In a RAG system, the prompt's context length, which includes chunk number and chunk window size, are relatively straightforward parameters to optimize. We developed the following \textbf{Algorithm~\ref{alg:context_length_adjustment}} to search these two parameters, implementing two different use cases:

\subsubsection{Threshold Method} In this approach, a predefined threshold is set. During execution, if the confidence score reaches or exceeds this threshold, the model proceeds to answer the question and exits the parameter search process.

\subsubsection{Best Probability Method} When time constraints are not a concern, this approach searches the entire parameter space to find the configuration that yields the highest probability, thereby maximizing accuracy under the given settings.

\begin{algorithm}[b] 
\caption{Hyperparameter Optimization}
\label{alg:context_length_adjustment}

\begin{algorithmic}[1]
\FOR{each $chunk\_number$ in the specified range}
    \FOR{each $window\_size$ in the specified range}
        \STATE Retrieve and generate prompt based on $chunk\_number$ and $window\_size$
        \STATE Predict answer and calculate first token probability
        \IF{using threshold method and confidence $> threshold$}
            \STATE Return predicted answer and exit
        \ELSIF{using best probability method and confidence $>$ previous best}
            \STATE Update best answer and confidence
        \ENDIF
    \ENDFOR
\ENDFOR
\STATE Return final answer based on the selected method
\end{algorithmic}
\end{algorithm}

\section{Experiments}
In this section, we first evaluated different RAG settings to determine the most effective settings. Building on these settings, we validated the earlier discussion on the relationship between first token probability and accuracy. We then conducted experiments on two distinct methods: the threshold method and the best probability method. Our results demonstrate that the highest accuracy achieved on the testset was 78.4\%.

Our evaluation is based on a dataset of 366 test samples, identical to the one provided by the challenge~\cite{ZindiAIML5G2024}, and is used to assess the model's accuracy.

\subsection{Settings of RAG}
In our experiments, we evaluated various configurations to optimize the performance of the RAG system. To mitigate the potential fragmentation of sentences, we employed the LlamaIndex SentenceSplitter, ensuring that the chunks retained coherence. We used a fixed chunk size of 64 tokens, which allowed for selective retrieval of relevant information from the 554 3GPP documents. Based on our calculations, this chunk size corresponds to an average of approximately 45 tokens when processed with the phi-2 model. To ensure that the total input does not exceed phi-2's context window limit of 2048 tokens, we restricted the number of chunks to no more than 35.

\begin{figure}[h]
    \centering
    \includegraphics[width=0.4\textwidth]{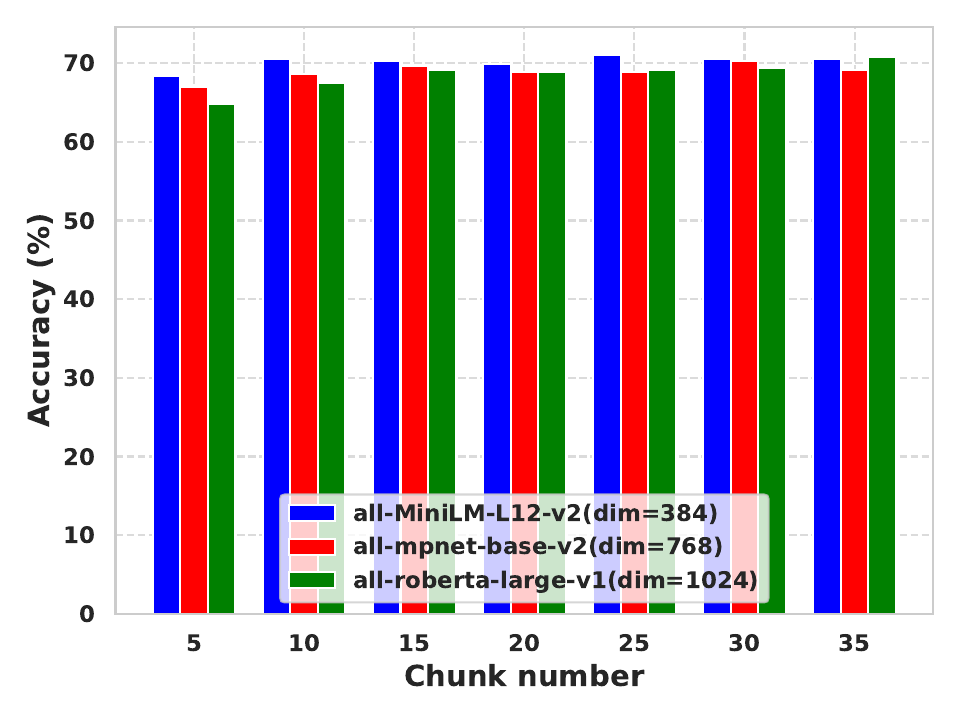}
    \caption{Accuracy versus the number of top K chunks included in the prompt, comparing the performance of three different embedding models.}
    \label{fig:top_k}
\end{figure}

We experimented with different top-K values, ranging from 5 to 35 in increments of 5, and set the window size to 0 to explore the impact on retrieval accuracy. For indexing, we utilized FAISS~\cite{douze2024faiss} with the IndexFlatIP method, ensuring efficient and accurate retrieval. The relationship between RAG accuracy and the top-K parameter is illustrated in Figure \ref{fig:top_k}, showing that the highest accuracy was not always achieved at the maximum $k$ value, likely due to the introduction of irrelevant chunks leading to comprehension disruption within the LLM. Additionally, our experiments revealed that embedding models with larger output dimensions did not necessarily result in better performance.

The settings and configurations used in our RAG system are summarized in Table \ref{tab:RAG_settings}. 

\begin{table}[h]
    \centering
    \caption{Settings and Configurations of the RAG System}
    \label{tab:RAG_settings}
    \begin{adjustbox}{width=0.4\textwidth}
    \begin{tabular}{|l|c|}
        \hline
        \textbf{Component}                & \textbf{Setting}    
        \\ \hline
        \textbf{LLM}               & Phi-2  
        \\ \hline
        \textbf{RAG corpus}               & 554 3GPP documents 
        \\ \hline
        \textbf{Chunk size}               & 64 tokens                                  \\ \hline
        \textbf{Window size}              & 0 to 1                                   \\ \hline
        \textbf{Top-K chunk number ($k$)}    & 5 to 35 (incremented by 5) \\ \hline
        
        \textbf{Embedding model}          & sentence-transformers/all-MiniLM-L12-v2     \\ \hline
        \textbf{Indexing strategy}        & IndexFlatIP                                \\ \hline
        \textbf{Sentence splitting}       & LlamaIndex's SentenceSplitter              \\ \hline
    \end{tabular}
    \end{adjustbox}
\end{table}

\subsection{Evaluating First Token Probability as a Confidence Indicator}

To validate the use of first token probability as a confidence indicator in our Confidence-Guided Context Adjustment mechanism, we conducted experiments on the testset with top 5-chunk RAG. 

\begin{figure}[h]
    \centering
    \includegraphics[width=0.4\textwidth]{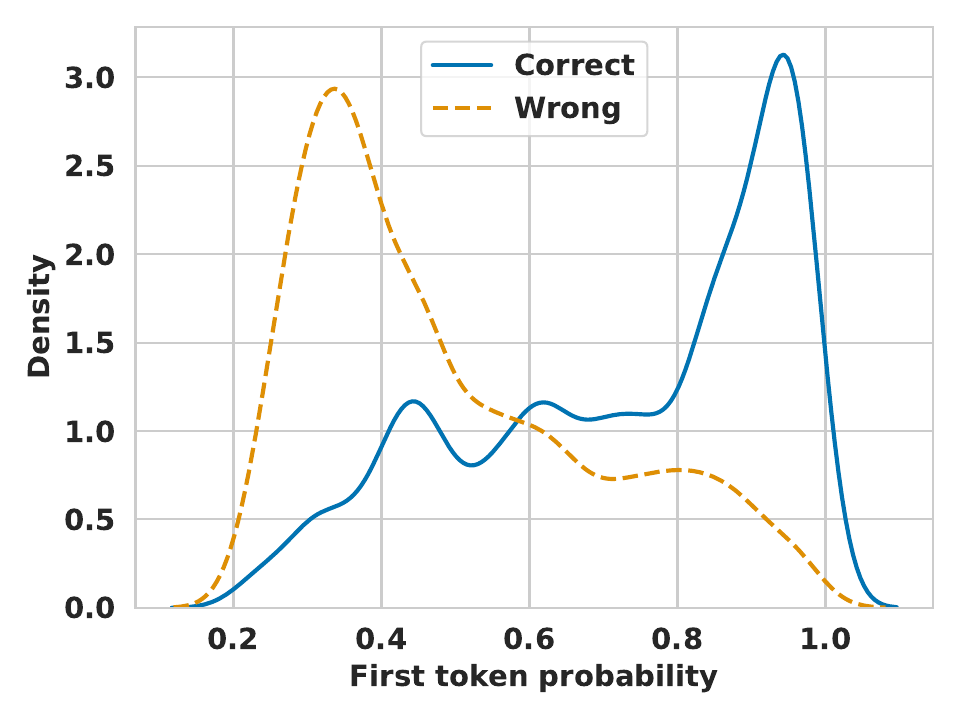}
    % \caption{Accuracy as a function of the cumulative context size, represented by the number of chunks included in the prompt. The accuracy is determined by selecting the option with the highest probability after multiple inference passes}
    \caption{Probability distributions of correct and wrong predictions with RAG. The x-axis denotes the first token probability, and the y-axis shows the normalized density of these predictions.}
    \label{fig:prob}
\end{figure}

We then separated the First Token Probabilities into two groups: one for correct answers and one for incorrect answers. As shown in Figure \ref{fig:prob}, correct answers generally have higher First Token Probabilities compared to incorrect ones, indicating that the model is more confident when it predicts correctly.

This clear difference in distributions confirms that lower First Token Probabilities are associated with lower accuracy, while higher First Token Probabilities correlate with better accuracy. This finding supports using first token probability as a reliable metric to guide context adjustments and improve the model's predictions.

\subsection{First Token Probability Guided RAG}

In this experiment, we aim to validate the effectiveness of the threshold method. Specifically, we investigate how many questions can be answered correctly under different chunk sizes when applying a fixed probability threshold, and assess the accuracy of those answers.

\begin{figure}[h]
    \centering
    \includegraphics[width=0.4\textwidth]{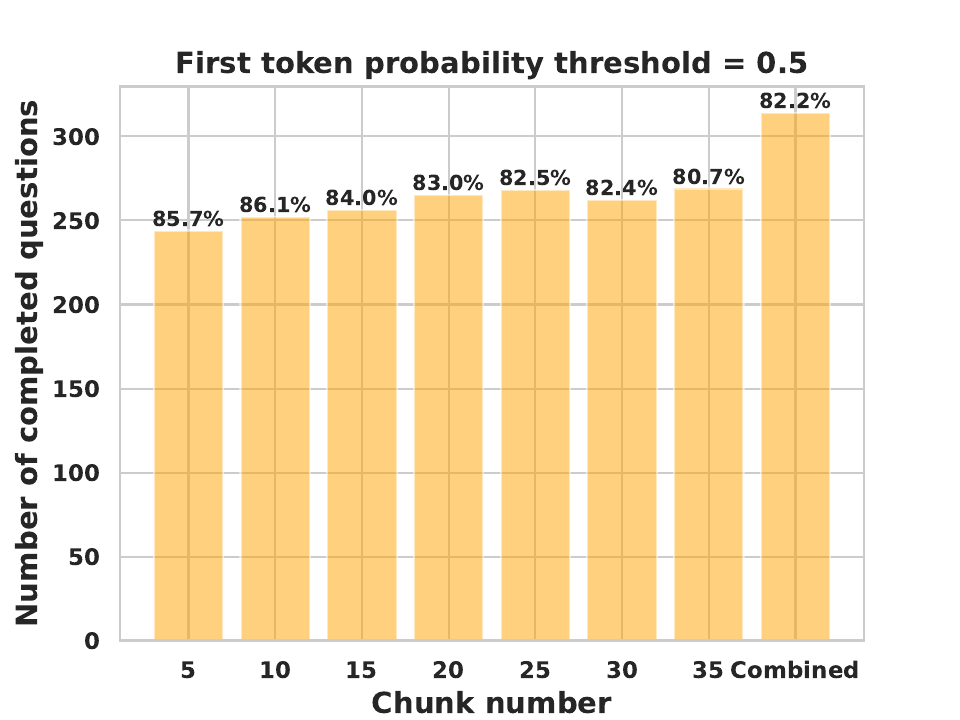}
    \caption{Number of completed questions and corresponding accuracy as a function of chunk number. The rightmost bar represents the results where the highest probability answer is selected from all chunk numbers combined.}

    \label{fig:Top1_question}
\end{figure}

Figure \ref{fig:Top1_question} illustrates the relationship between accuracy and chunk number, with the window size set to 0, while other settings remain consistent with those in Table \ref{tab:RAG_settings}. The accuracy is calculated by selecting the option with the highest probability after multiple inference passes.

The results show that by setting a threshold of 0.5, the model can correctly answer approximately 250 questions, all with an accuracy exceeding 80\%. Furthermore, when searching across all chunk sizes and selecting the token with the highest probability, the model is able to answer 315 questions.

\subsection{Maximizing Accuracy through Best Probability Method}

In this experiment, we evaluate the model's accuracy by running the model across various chunk sizes, each combined with two different context windows. For each chunk and window configuration, the model generates a probability distribution over possible answers. The final answer is selected based on the highest probability observed across all configurations, which we refer to as the "Combined" result.

This approach allows us to explore how combining different context windows and chunk sizes can impact the model's performance. Notably, as chunk size increases, especially under window 1 configurations, the model may experience truncation due to context length limits. This truncation can affect the accuracy for larger chunk sizes.

Notably, the results presented in Table \ref{table:2} reveal a significant trend: as more results from different contexts, chunk sizes, and even embedding models are combined, the model's accuracy continues to improve. The combined results from both context windows show a significant accuracy improvement. However, when we integrate results across different embedding models, the accuracy further increases, leading to a total improvement of 26.8\% over the baseline without RAG. This demonstrates the advantage of leveraging diverse configurations, with the combined embedding models contributing substantially to the overall accuracy gain.

\begin{table}[h]
    \centering
    \caption{Accuracy for Different Chunk Numbers and Window Sizes. The results show the accuracy for each chunk number under two different window configurations (Window 0 and Window 1). The table also includes combined results across both window configurations, combined embedding models, and the baseline accuracy without RAG.}

    \begin{tabular}{c|c|c}
        \hline
        \textbf{Chunk number} & \textbf{Window 0} & \textbf{Window 1} \\
        \hline
        5  & 68.3\% & 72.1\% \\
        10 & 70.49\% & 74.8\% \\
        15 & 70.2\% & 70.7\% \\
        20 & 69.9\% & 70.4\% \\
        25 & 71.0\% & 70.7\% \\
        30 & 70.4\% & 68.0\% \\
        35 & 70.4\% & 68.0\% \\
        \hline
        \textbf{Combined} & 74.5\% & 76.7\% \\
        \hline
        \textbf{Combined window} & \multicolumn{2}{c}{77.6\%} \\
        \hline
        \textbf{Combined embeddings models} & \multicolumn{2}{c}{78.4\%} \\
        \hline
        \textbf{Without RAG} & \multicolumn{2}{c}{51.6\%} \\
        \hline
    \end{tabular}
    \label{table:2}
\end{table}

% \begin{figure}[h]
%     \centering
%     \includegraphics[width=0.5\textwidth]{img/accuracy_vs_combined_configurations.pdf}
%     \caption{Accuracy as a function of the cumulative context size, represented by the number of chunks included in the prompt. The accuracy is determined by selecting the option with the highest probability after multiple inference passes}
%     \label{fig:result}
% \end{figure}

\section{Conclusion}

In this paper, we introduced a novel approach to enhance the performance of LLMs in domain-specific MCQA tasks. By leveraging RAG and first token probability guided chunk number and window size optimization, we demonstrated that it is possible to significantly improve the accuracy and reliability of model predictions. Our experiments validated the effectiveness of the proposed methods, showing a clear correlation between first token probability and answer's accuracy. Furthermore, the use of RAG allowed for the integration of domain-specific knowledge without the need for extensive model retraining. These findings underscore the potential of irst token probability guided RAG in overcoming the inherent challenges associated with LLMs, such as hallucinations and limited explainability. In future work, we plan to explore further optimizations and extend this approach to other specialized domains, as well as investigate the impact of LLM-generated token probability on performance.

\bibliographystyle{IEEEtran}
\bibliography{ref}

\end{document}